\documentclass[11pt,a4paper]{article}
\usepackage[hyperref]{conll2018}
\usepackage{times}
\usepackage{latexsym}
\usepackage{graphicx}
\usepackage{float}
\usepackage{amsmath}
\usepackage{amssymb}
\usepackage{multirow}
\usepackage{booktabs}
\usepackage{url}
\usepackage{array}
\usepackage{enumitem}

\aclfinalcopy 

\title{Contextual Recurrent Units for Cloze-style Reading Comprehension}

\author{Yiming Cui$^{\dag\ddag}$, Wei-Nan Zhang$^\dag$, Wanxiang Che$^\dag$, Ting Liu$^\dag$, \\
{\bf Zhipeng Chen$^\ddag$, Shijin Wang$^\ddag$$^\S$, Guoping Hu$^\ddag$ } \\
{$^\dag$Research Center for Social Computing and Information Retrieval (SCIR),}\\
{Harbin Institute of Technology, Harbin, China}\\
{$^\ddag$State Key Laboratory of Cognitive Intelligence, iFLYTEK Research, China}\\
{$^\S$iFLYTEK AI Research (Hebei), Langfang, China} \\
{$^\dag$\tt \{ymcui,wnzhang,car,tliu\}@ir.hit.edu.cn}\\
{$^\ddag$$^\S$\tt\{ymcui,zpchen,sjwang3,gphu\}@iflytek.com}\\  
}

\date{}

\begin{document}
\maketitle
\begin{abstract}
Recurrent Neural Networks (RNN) are known as powerful models for handling sequential data,
and especially widely utilized in various natural language processing tasks.
In this paper, we propose Contextual Recurrent Units (CRU) for enhancing local contextual representations in neural networks.
The proposed CRU injects convolutional neural networks (CNN) into the recurrent units to enhance the ability to model the local context and reducing word ambiguities even in bi-directional RNNs.
We tested our CRU model on sentence-level and document-level modeling NLP tasks: sentiment classification and reading comprehension.
Experimental results show that the proposed CRU model could give significant improvements over traditional CNN or RNN models, including bidirectional conditions, as well as various state-of-the-art systems on both tasks, showing its promising future of extensibility to other NLP tasks as well.
\end{abstract}

\section{Introduction}\label{introduction}
Neural network based approaches have become popular frameworks in many machine learning research fields, showing its advantages over traditional methods. In NLP tasks, two types of neural networks are widely used: Recurrent Neural Network (RNN) and Convolutional Neural Network (CNN). 

RNNs are powerful models in various NLP tasks, such as machine translation \citep{cho-etal-2014}, sentiment classification \citep{wang-and-tian-2016,liu-emnlp-2016,wang-etal-2016,zhang-etal-2016,liang-etal-2016}, reading comprehension \citep{kadlec-etal-2016,dhingra-etal-2016,sordoni-etal-2016,cui-etal-2016,cui-etal-2017-aoa,yang-etal-2016}, etc. 
The recurrent neural networks can flexibly model different lengths of sequences into a fixed representation. 
There are two main implementations of RNN: Long Short-Term Memory (LSTM) \citep{hochreiter-1997} and Gated Recurrent Unit (GRU) \citep{cho-etal-2014}, which solve the gradient vanishing problems in vanilla RNNs. 

Compared to RNN, the CNN model also shows competitive performances in some tasks, such as text classification \citep{kim-2014}, etc.
However, different from RNN, CNN sets a pre-defined convolutional kernel to ``summarize'' a fixed window of adjacent elements into blended representations, showing its ability of modeling local context.

As both global and local information is important in most of NLP tasks \citep{luong-etal-2015}, in this paper, we propose a novel recurrent unit, called Contextual Recurrent Unit (CRU). The proposed CRU model adopts advantages of RNN and CNN, where CNN is good at modeling local context, and RNN is superior in capturing long-term dependencies. We propose three variants of our CRU model: {\em shallow fusion}, {\em deep fusion} and {\em deep-enhanced fusion}. 

To verify the effectiveness of our CRU model, we utilize it into two different NLP tasks: sentiment classification and reading comprehension, where the former is sentence-level modeling, and the latter is document-level modeling. 
In the sentiment classification task, we build a standard neural network and replace the recurrent unit by our CRU model.
To further demonstrate the effectiveness of our model, we also tested our CRU in reading comprehension tasks with a strengthened baseline system originated from Attention-over-Attention Reader (AoA Reader) \citep{cui-etal-2017-aoa}.
Experimental results on public datasets show that our CRU model could substantially outperform various systems by a large margin, and set up new state-of-the-art performances on related datasets.
The main contributions of our work are listed as follows.
\begin{itemize}[leftmargin=*]
	\item We propose a novel neural recurrent unit called Contextual Recurrent Unit (CRU), which effectively incorporate the advantage of CNN and RNN. Different from previous works, our CRU model shows its excellent flexibility as GRU and provides better performance.
	\item The CRU model is applied to both sentence-level and document-level modeling tasks and gives state-of-the-art performances.
	\item The CRU could also give substantial improvements in cloze-style reading comprehension task when the baseline system is strengthened by incorporating additional features which will enrich the representations of unknown words and make the texts more readable to the machine.
\end{itemize}

\section{Related Works}\label{related-work}

Gated recurrent unit (GRU) has been proposed in the scenario of neural machine translations \citep{cho-etal-2014}. It has been shown that the GRU has comparable performance in some tasks compared to the LSTM. Another advantage of GRU is that it has a simpler neural architecture than LSTM, showing a much efficient computation.

However, convolutional neural network (CNN) is not as popular as RNNs in NLP tasks, as the texts are formed temporally. But in some studies, CNN shows competitive performance to the RNN models, such as text classification \citep{kim-2014}.

Various efforts have been made on combining CNN and RNN.
\citet{wang-etal-2016} proposed an architecture that combines CNN and GRU model with pre-trained word embeddings by word2vec. 
\citet{liang-etal-2016} proposed to combine asymmetric convolution neural network with the bidirectional LSTM network. 
\citet{zhang-etal-2016} presented Dependency Sensitive CNN, which hierarchically construct text by using LSTMs and extracting features with convolution operations subsequently. 
\citet{cai-etal-2016} propose to make use of dependency relations information in the shortest dependency path (SDP) by combining CNN and two-channel LSTM units. 
\citet{kim-etal-2016} build a neural network for dialogue topic tracking where the CNN used to account for semantics at individual utterance and RNN for modeling conversational contexts along multiple turns in history.

The difference between our CRU model and previous works can be concluded as follows.
\begin{itemize}[leftmargin=*]
  \item Our CRU model could adaptively control the amount of information that flows into different gates, which was not studied in previous works.
  \item Also, the CRU does not introduce a pooling operation, as opposed to other works, such as CNN-GRU \citep{wang-etal-2016}. Our motivation is to provide flexibility as the original GRU, while the pooling operation breaks this law (the output length is changed), and it is unable to do exact word-level attention over the output. However, in our CRU model, the output length is the same as the input's and can be easily applied to various tasks where the GRU used to. 
  \item We also observed that by only using CNN to conclude contextual information is not strong enough. So we incorporate the original word embeddings to form a "word + context" representation for enhancement.
\end{itemize}

\section{Our approach}\label{cru}
In this section, we will give a detailed introduction to our CRU model.
Firstly, we will give a brief introduction to GRU \citep{cho-etal-2014} as preliminaries, and then three variants of our CRU model will be illustrated.

\subsection{Gated Recurrent Unit}
Gated Recurrent Unit (GRU) is a type of recurrent unit that models sequential data \citep{cho-etal-2014}, which is similar to LSTM but is much simpler and computationally effective than the latter one. We will briefly introduce the formulation of GRU.
Given a sequence $x = \{x_1, x_2, ..., x_n\}$, GRU will process the data in the following ways. For simplicity, the bias term is omitted in the following equations.
\begin{gather}
z_t = \sigma(W_z x_t+U_z h_{t-1}) \\
r_t = \sigma(W_r x_t+U_r h_{t-1}) \\
\widetilde{h_t} = \tanh(W x_t+U [r_t \odot h_{t-1}]) \\
h_t = z_t  h_{t-1} + (1-z_t) \widetilde{h_t}
\end{gather}

where $z_t$ is the update gate, $r_t$ is the reset gate, and non-linear function $\sigma$ is often chosen as $sigmoid$ function.
In many NLP tasks, we often use a bi-directional GRU, which takes both forward and backward information into account.

\subsection{Contextual Recurrent Unit}
By only modeling word-level representation may have drawbacks in representing the word that has different meanings when the context varies. 
Here is an example that shows this problem.

\begin{quote}
\begin{scriptsize}\begin{verbatim}
There are many fan mails in the mailbox. 
There are many fan makers in the factory.
\end{verbatim}\end{scriptsize}
\end{quote}

As we can see that, though two sentences share the same beginning before the word {\em fan}, the meanings of the word {\em fan} itself are totally different when we meet the following word {\em mails} and {\em makers}. The first {\em fan} means ``a person that has strong interests in a person or thing", and the second one means ``a machine with rotating blades for ventilation".
However, the embedding of word {\em fan} does not discriminate according to the context. 
Also, as two sentences have the same beginning, when we apply a recurrent operation (such as GRU) till the word {\em fan}, the output of GRU does not change, though they have entirely different meanings when we see the following words.

To enrich the word representation with local contextual information and diminishing the word ambiguities, we propose a model as an extension to the GRU, called Contextual Recurrent Unit (CRU).
In this model, we take full advantage of the convolutional neural network and recurrent neural network, where the former is good at modeling local information, and the latter is capable of capturing long-term dependencies.
Moreover, in the experiment part, we will also show that our bidirectional CRU could also significantly outperform the bidirectional GRU model.

In this paper, we propose three different types of CRU models: {\em shallow fusion}, {\em deep fusion} and {\em deep-enhanced fusion},  from the most fundamental one to the most expressive one. 
We will describe these models in detail in the following sections.

\subsubsection{Shallow Fusion}
The most simple one is to directly apply a CNN layer after the embedding layer to obtain blended contextual representations. Then a GRU layer is applied afterward. We call this model as {\em shallow fusion}, because the CNN and RNN are applied linearly without changing inner architectures of both. 

Formally, when given a sequential data $x = \{x_1, x_2, ..., x_n\}$, a shallow fusion of CRU can be illustrated as follows.
\begin{gather} 
e_t = W_e \cdot x_t ~~;~~ c_t = \phi(\widetilde{e_t}) \\
h_t = GRU(h_{t-1}, c_t) 
\end{gather}

We first transform word $x_t$ into word embeddings through an embedding matrix $W_e$.
Then a convolutional operation $\phi$ is applied to the context of $e_t$, denoted as $\widetilde{e_t}$, to obtain contextual representations.
Finally, the contextual representation $c_t$ is fed into GRU units.

Following \cite{kim-2014}, we apply embedding-wise convolution operation, which is commonly used in natural language processing tasks.
Let $e_{i:j} \in\mathbb{R}^{\mathcal \\j*d}$ denote the concatenation of $j-i+1$ consecutive $d$-dimensional word embeddings.
\begin{equation} e_{i:j} = concat[e_i, e_{i+1}, ..., e_j] \end{equation}

The embedding-wise convolution is to apply a convolution filter {\bf w} $\in\mathbb{R}^{\mathcal \\k*d}$  to a window of $k$ word embeddings to generate a new feature, i.e., summarizing a local context of $k$ words. 
This can be formulated as
\begin{equation} c_i = f({\bf w} \cdot e_{i:i+k-1} + b) \end{equation}
where $f$ is a non-linear function and $b$ is the bias.

By applying the convolutional filter to all possible windows in the sentence, a feature map $c$ will be generated.
In this paper, we apply a {\em same-length} convolution (length of the sentence does not change), i.e. $c \in\mathbb{R}^{\mathcal \\n*1}$.
Then we apply $d$ filters with the same window size to obtain multiple feature maps.
So the final output of CNN has the shape of $C \in\mathbb{R}^{\mathcal \\n*d}$, which is exactly the same size as $n$ word embeddings, which enables us to do exact word-level attention in various tasks.

\subsubsection{Deep Fusion}
The contextual information that flows into the update gate and reset gate of GRU is identical in shallow fusion.
In order to let the model adaptively control the amount of information that flows into these gates, we can embed CNN into GRU in a deep manner. We can rewrite the Equation 1 to 3 of GRU as follows.
\begin{gather}
z_t = \sigma(\phi_z(\widetilde{e_t})) + U_z h_{t-1}) \\
r_t = \sigma(\phi_r(\widetilde{e_t})) + U_r h_{t-1}) \\
\widetilde{h_t} = \tanh(\phi(\widetilde{e_t}))+U [r_t \odot h_{t-1}]) 
\end{gather}

where $\phi_z, \phi_r, \phi$ are three different CNN layers, i.e., the weights are not shared.
When the weights share across these CNNs, the deep fusion will be degraded to shallow fusion.

\subsubsection{Deep-Enhanced Fusion}
In shallow fusion and deep fusion, we used the convolutional operation to summarize the context.
However, one drawback of them is that the original word embedding might be blurred by blending the words around it, i.e., applying the convolutional operation on its context. 

\begin{figure}[tp]
  \centering
  \includegraphics[width=0.5\textwidth]{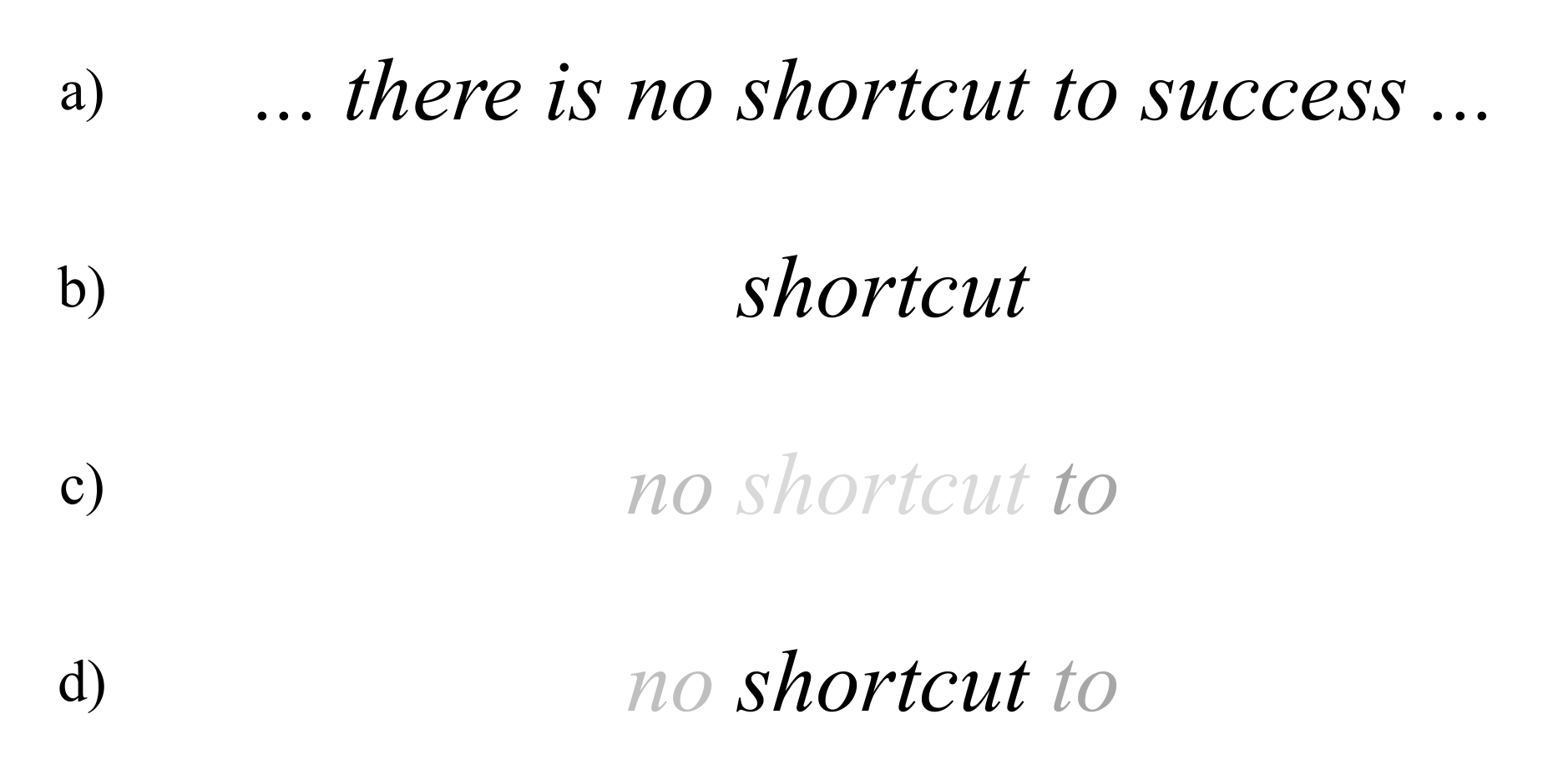}
  \caption{\label{deep-e-example} An intuitive illustration of variants of the CRU model. The gray scale represents the amount of information. a) original sentence; b) original representation of word ``shortcut''; c) applying convolutional filter (length=3); d) adding original word embedding; }
\end{figure}

For better modeling the original word and its context, we enhanced the deep fusion model with original word embedding information, with an intuition of ``enriching word representation with contextual information while preserving its basic meaning''.
Figure \ref{deep-e-example} illustrates our motivations.

Formally, the Equation 9 to 11 can be further rewritten into
\begin{gather}
z_t = \sigma(W_z(\phi_z(\widetilde{e_t}) + e_t) + U_z h_{t-1} \\
r_t = \sigma(W_r(\phi_r(\widetilde{e_t}) + e_t) + U_r h_{t-1}) \\
\widetilde{h_t} = \tanh(W(\phi(\widetilde{e_t}) + e_t)+U [r_t \odot h_{t-1}])
\end{gather}

where we add original word embedding $e_t$ after the CNN operation, to ``enhance'' the original word information while not losing the contextual information that has learned from CNNs.

\section{Applications}\label{application}
The proposed CRU model is a general neural recurrent unit, so we could apply it to various NLP tasks.
As we wonder whether the CRU model could give improvements in both sentence-level modeling and document-level modeling tasks, in this paper, we applied the CRU model to two NLP tasks: sentiment classification and cloze-style reading comprehension.
In the sentiment classification task, we build a simple neural model and applied our CRU.
In the cloze-style reading comprehension task, we first present some modifications to a recent reading comprehension model, called AoA Reader \cite{cui-etal-2017-aoa}, and then replace the GRU part by our CRU model to see if our model could give substantial improvements over strong baselines.

\subsection{Sentiment Classification}\label{sentiment-classification}
\begin{figure}[tp]
  \centering
  \includegraphics[width=0.48\textwidth]{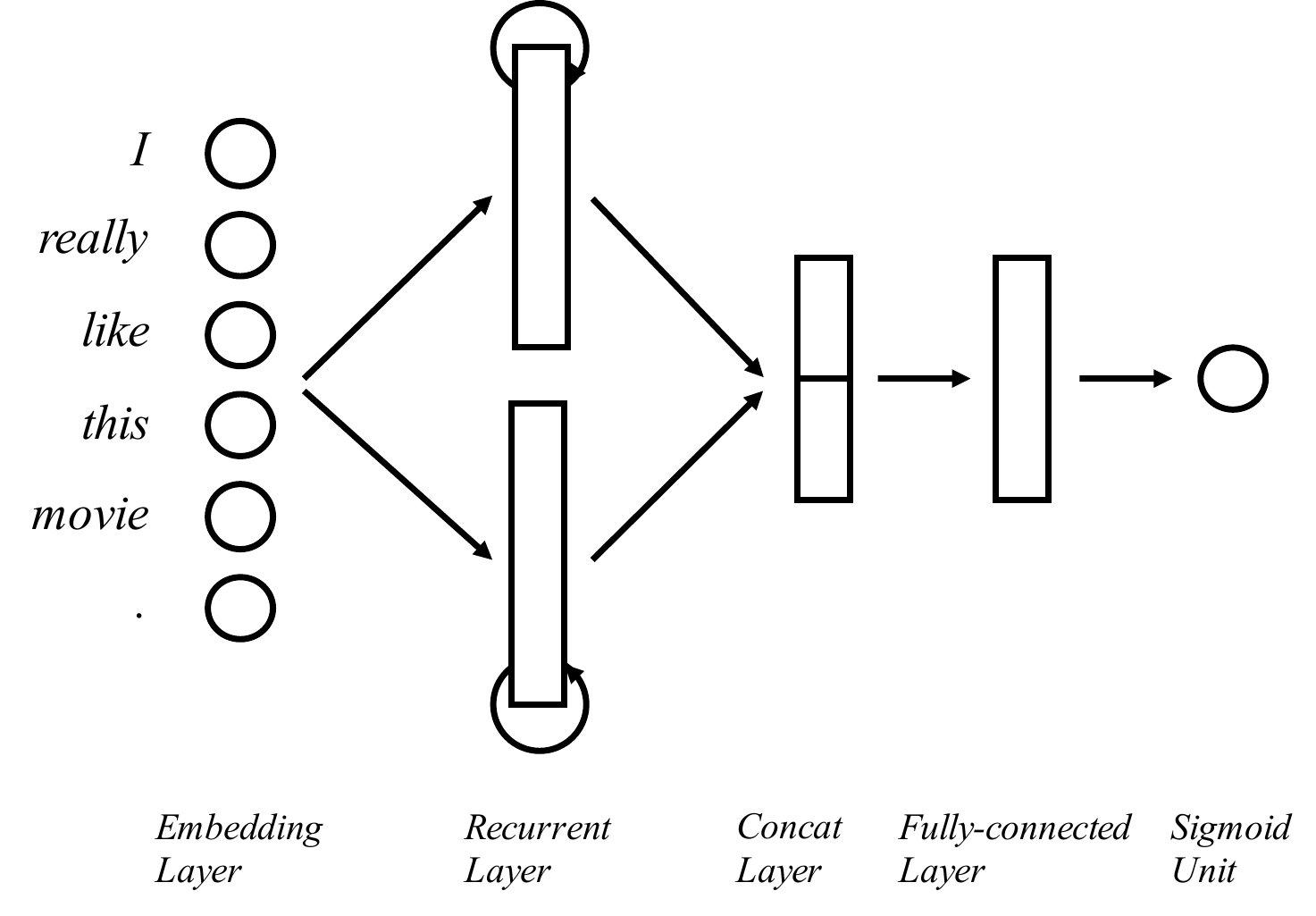}
  \caption{\label{sc-arch} A general neural network architecture of sentiment classification task.}
\end{figure}

In the sentiment classification task, we aim to classify movie reviews, where one movie review will be classified into the positive/negative or subjective/objective category.
A general neural network architecture for this task is depicted in Figure \ref{sc-arch}.

First, the movie review is transformed into word embeddings.
And then, a sequence modeling module is applied, in which we can adopt LSTM, GRU, or our CRU, to capture the inner relations of the text.
In this paper, we adopt bidirectional recurrent units for modeling sentences, and then the final hidden outputs are concatenated.
After that, a fully connected layer will be added after sequence modeling.
Finally, the binary decision is made through a single $sigmoid$ unit.

As shown, we employed a straightforward neural architecture to this task, as we purely want to compare our CRU model against other sequential models.
The detailed experimental result of sentiment classification will be given in the next section.

\subsection{Reading Comprehension}\label{rc-task}
Besides the sentiment classification task, we also tried our CRU model in cloze-style reading comprehension, which is a much complicated task. 
In this paper, we strengthened the recent AoA Reader \cite{cui-etal-2017-aoa} and applied our CRU model to see if we could obtain substantial improvements when the baseline is strengthened.

\subsubsection{Task Description}
The cloze-style reading comprehension is a fundamental task that explores relations between the document and the query.
Formally, a general cloze-style query can be illustrated as a triple $\langle {\mathcal D}, {\mathcal Q}, {\mathcal A} \rangle$, where $\mathcal D$ is the document, $\mathcal Q$ is the query and the answer $\mathcal A$. 
Note that the answer is a {\em single} word in the document, which requires us to exploit the relationship between the document and query.

\subsubsection{Modified AoA Reader}
In this section, we briefly introduce the original AoA Reader \cite{cui-etal-2017-aoa}, and illustrate our modifications.
When a cloze-style training triple $\langle \mathcal D, \mathcal Q, \mathcal A \rangle$ is given, the Modified AoA Reader will be constructed in the following steps.
First, the document and query will be transformed into continuous representations with the embedding layer and recurrent layer.
The recurrent layer can be the simple RNN, GRU, LSTM, or our CRU model.

To further strengthen the representation power, we show a simple modification in the embedding layer, where we found strong empirical results in performance.
The main idea is to utilize additional sparse features of the word and add (concatenate) these features to the word embeddings to enrich the word representations. The additional features have shown effective in various models \cite{dhingra-etal-2016,pengli-etal-2016,yang-etal-2016}.
In this paper, we adopt two additional features in document word embeddings (no features applied to the query side).

\noindent{{$\bullet$~~ \bf Document word frequency}}: Calculate each document word frequency. This helps the model to pay more attention to the important (more mentioned) part of the document.
\begin{equation} freq(d) = \frac{word\_count(d)}{length(\mathcal D)}, d\in \mathcal D \end{equation}

\noindent{{$\bullet$~~ \bf Count of query word}}: Count the number of each document word appeared in the query. For example, if a document word appears three times in the query, then the feature value will be 3. We empirically find that instead of using binary features (appear=1, otherwise=0) \cite{pengli-etal-2016}, indicating the count of the word provides more information, suggesting that the more a word occurs in the query, the less possible the answer it will be.
We replace the Equation 16 with the following formulation (query side is not changed), 
\begin{equation}  \small e(x) = concat[W_e \cdot x, freq(x), CoQ(x)] , x\in \mathcal D \end{equation}
where $freq(x)$ and $CoQ(x)$ are the features that introduced above. 
\begin{gather}
{\small \overrightarrow{h_s(x)}} =  {\small \overrightarrow{RNN}(e(x)) ; \overleftarrow{h_s(x)} = \overleftarrow{RNN}(e(x))} \\
h_s(x) = [\overrightarrow{h_s(x)}; \overleftarrow{h_s(x)}]
\end{gather}

Other parts of the model remain the same as the original AoA Reader. For simplicity, we will omit this part, and the detailed illustrations can be found in \citet{cui-etal-2017-aoa}.

\section{Experiments: Sentiment Classification}\label{experiments-sc}

\subsection{Experimental Setups}
In the sentiment classification task, we tried our model on the following public datasets.
\begin{itemize}[leftmargin=*]
  \item {\bf MR}\footnote{\url{http://www.cs.cornell.edu/People/pabo/movie-review-data/}} Movie reviews with one sentence each. Each review is classified into positive or negative \cite{pang-and-lee-2005}.
  \item {\bf IMDB}\footnote{\url{http://ai.stanford.edu/~amaas/data/sentiment/}} Movie reviews from IMDB website, where each movie review is labeled with binary classes, either positive or negative \cite{maas-etal-2011}. Note that each movie review may contain several sentences.
  \item {\bf SUBJ}$^1$ Movie review labeled with subjective or objective \cite{pang-and-lee-2004}. 
\end{itemize}

The statistics and hyper-parameter settings of these datasets are listed in Table \ref{imdb-stats}.
 
        \begin{table}[htp]
        \begin{center}
        \begin{tabular}{lccc}
        \toprule
        & \bf MR & \bf IMDB & \bf SUBJ \\
        \midrule
        Train \# & 10,662 & 25,000 & 10,000 \\
        Test \# & 10-CV & 25,000 & 10-CV \\
        \midrule
        Embed. size & 200 & 256 & 200 \\
        Hidden size & 200 & 256 & 200 \\
        Dropout & 0.3 & 0.3 & 0.4 \\
        Pre-train Embed. & GloVe & - & GloVe \\
        Initial LR & 0.0005 & 0.001 & 0.0005 \\
        Vocab truncation & - & 50,000 & - \\
        \bottomrule
        \end{tabular}
        \end{center}
        \caption{\label{imdb-stats} Statistics and hyper-parameter settings of MR, IMDB and SUBJ datasets. 10-CV represents 10-fold cross validation.}
        \end{table}

As these datasets are quite small and overfit easily, we employed $l_2$-regularization of 0.0001 to the embedding layer in all datasets. 
Also, we applied dropout \cite{srivastava-etal-2014} to the output of the embedding layer and fully connected layer.
The fully connected layer has a dimension of 1024.
In the MR and SUBJ, the embedding layer is initialized with 200-dimensional GloVe embeddings (trained on 840B token) \cite{pennington-etal-2014} and fine-tuned during the training process.
In the IMDB condition, the vocabulary is truncated by descending word frequency order.
We adopt batched training strategy of 32 samples with ADAM optimizer \cite{kingma2014adam}, and clipped gradient to 5 \cite{pascanu-etal-2013}.
Unless indicated, the convolutional filter length is set to 3, and ReLU for the non-linear function of CNN in all experiments.
We use 10-fold cross-validation (CV) in the dataset that has no train/valid/test division.     

\subsection{Results}\label{result-sc}

The experimental results are shown in Table \ref{exp-class}.
As we mentioned before, all RNNs in these models are {\bf bi-directional}, because we wonder if our bi-CRU could still give substantial improvements over bi-GRU which could capture both history and future information.
As we can see that, all variants of our CRU model could give substantial improvements over the traditional GRU model, where a maximum gain of 2.7\%, 1.0\%, and 1.9\% can be observed in three datasets, respectively.
We also found that though we adopt a straightforward classification model, our CRU model could outperform the state-of-the-art systems by 0.6\%, 0.7\%, and 0.8\% gains respectively, which demonstrate its effectiveness. 
By employing more sophisticated architecture or introducing task-specific features, we think there is still much room for further improvements, which is beyond the scope of this paper.

       \begin{table}[t]
        \begin{center}
        \small
        \begin{tabular}{lccc}
        \toprule
        \bf System & \bf MR & \bf IMDB & \bf SUBJ \\
        \midrule
        Multi-channel CNN & 81.1 & - & 93.2 \\
        HRL  & - & 90.9 & - \\
        Multi-task arc-II  & - & {\em 91.2} & {\em 95.0} \\
        CNN-GRU-wordvec  & 82.3 & - & - \\
        DSCNN-Pretrain  & 82.2 & 90.7 & 93.9 \\
        LR-Bi-LSTM & 82.1 & - & - \\
        AC-BLSTM & 83.1& - & 94.2 \\
        G-AC-BLSTM & {\em 83.7} & -& 94.3 \\
        \midrule\midrule
        GRU & 81.0 & 90.9 & 93.9 \\
        CRU (shallow fusion) & 82.1 & 91.3 & 95.0 \\
        CRU (deep fusion) & 82.7 & 91.5 & 95.2 \\
        CRU (deep-enhanced, filter=3) & {\bf 83.7} & {\bf 91.9} & {\bf 95.8} \\
        CRU (deep-enhanced, filter=5) & 83.2 & 91.7 & 95.2 \\
        \bottomrule
        \end{tabular}
        \end{center}
        \caption{\label{exp-class} Results on MR, IMDB and SUBJ sentiment classification task. Best previous results are marked in italics, and overall best results are mark in bold face. {\small {\bf Multi-channel CNN} \cite{kim-2014}: A CNN architecture with static and non-static word embeddings. {\bf HRL} \cite{wang-and-tian-2016}: A hybrid residual LSTM architecture. {\bf Multi-task arc-II} \cite{liu-emnlp-2016}: A deep architectures with shared local-global hybrid memory for multi-task learning. {\bf CNN-GRU-word2vec} \cite{wang-etal-2016}: An architecture that combines CNN and GRU model with pre-trained word embeddings by {\em word2vec}. {\bf DSCNN-Pretrain} \cite{zhang-etal-2016}: Dependency sensitive convolutional neural networks with pretrained sequence autoencoders. {\bf AC-BLSTM} \cite{liang-etal-2016}: Asymmetric convolutional bidirectional LSTM networks. } }
        \end{table}
        
\begin{figure}[t]
      \centering
      \includegraphics[width=0.45\textwidth]{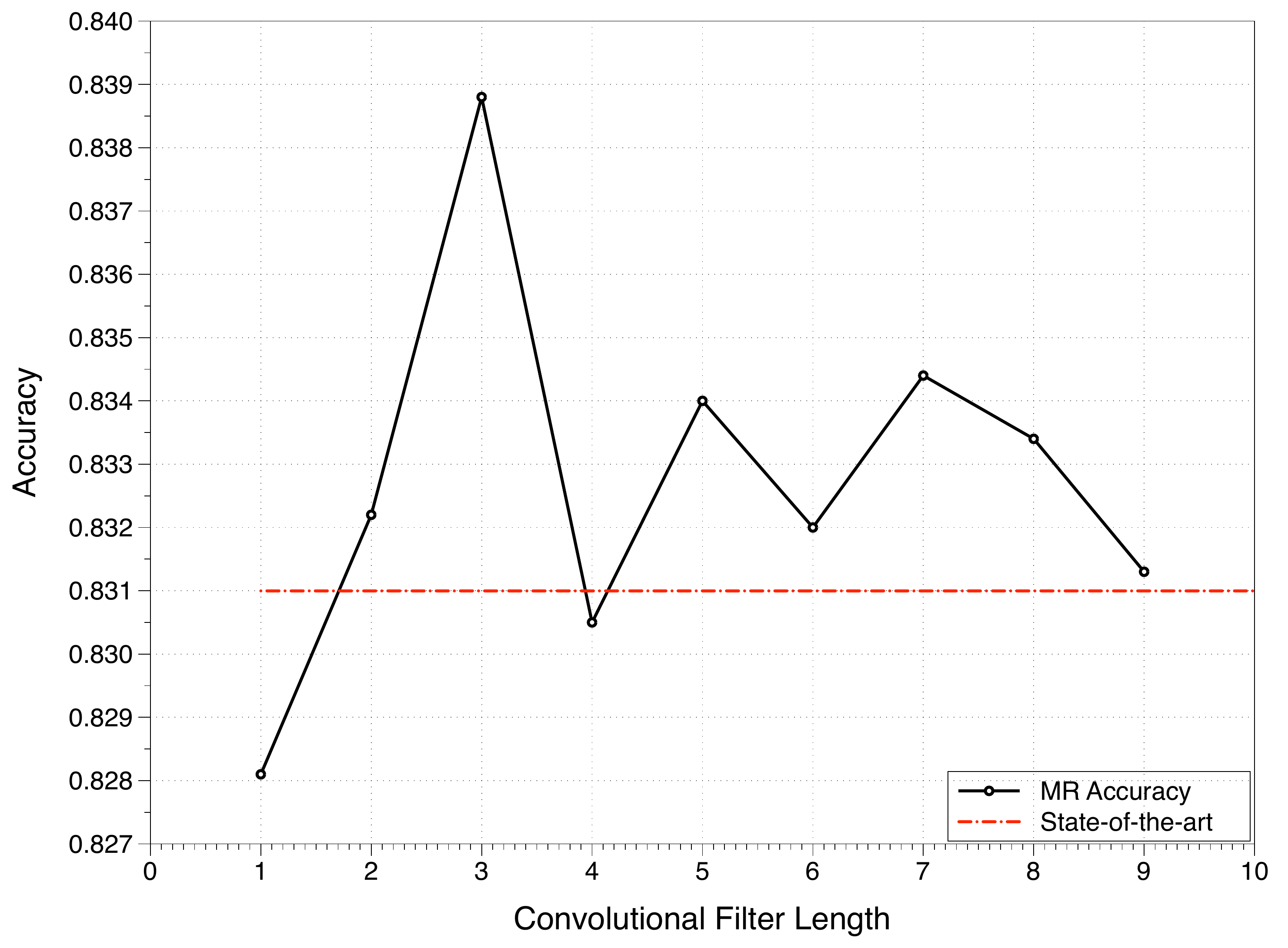}
      \caption{\label{mr-length} Trends of MR test set accuracy with the increasing convolutional filter length. }
    \end{figure}

When comparing three variants of the CRU model, as we expected, the CRU with {\em deep-enhanced fusion} performs best among them. This demonstrates that by incorporating contextual representations with original word embedding could enhance the representation power.
Also, we noticed that when we tried a larger window size of the convolutional filter, i.e., 5 in this experiment, does not give a rise in the performance. 
We plot the trends of MR test set accuracy with the increasing convolutional filter length, as shown in Figure \ref{mr-length}.

As we can see that, using a smaller convolutional filter does not provide much contextual information, thus giving a lower accuracy.
On the contrary, the larger filters generally outperform the lower ones, but not always.
One possible reason for this is that when the filter becomes larger, the amortized contextual information is less than a smaller filter, and make it harder for the model to learn the contextual information. However, we think the proper size of the convolutional filter may vary task by task. Some tasks that require long-span contextual information may benefit from a larger filter.
    
We also compared our CRU model with related works that combine CNN and RNN \cite{wang-etal-2016,zhang-etal-2016,liang-etal-2016}. 
From the results, we can see that our CRU model significantly outperforms previous works, which demonstrates that by employing {\em deep fusion} and enhancing the contextual representations with original embeddings could substantially improve the power of word representations.
    
\begin{figure}[tb]
      \centering
      \includegraphics[width=0.45\textwidth]{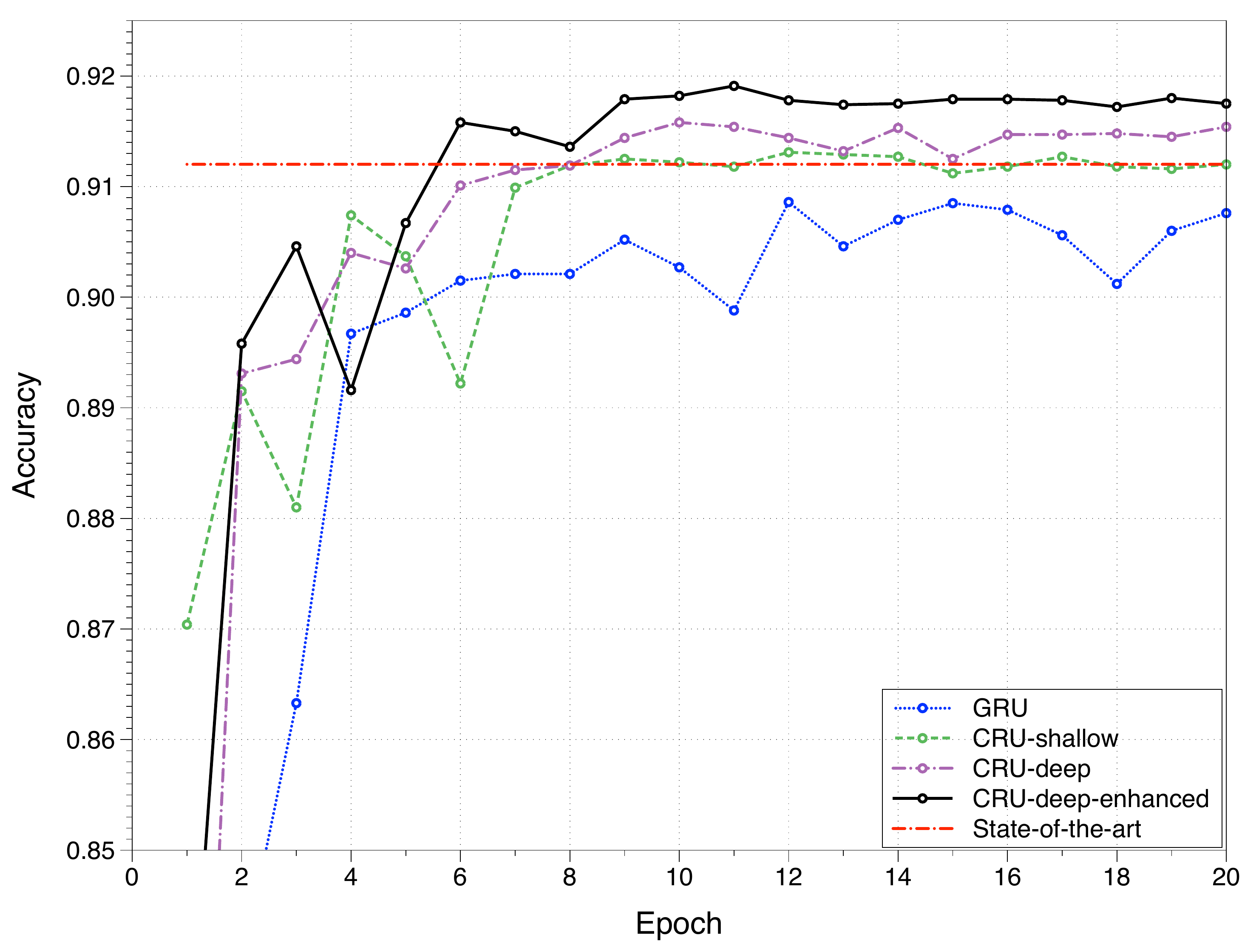}
      \caption{\label{imdb-train} Trends of IMDB test set accuracy with the training time growing.}
    \end{figure}

On another aspect, we plot the trends of IMDB test set accuracy during the training process, as depicted in Figure \ref{imdb-train}.
As we can see that, after iterating six epochs of training data, all variants of CRU models show faster convergence speed and smaller performance fluctuation than the traditional GRU model, which demonstrates that the proposed CRU model has better training stability.

\section{Experiments: Reading Comprehension}\label{experiments-rc}        
    
    \begin{table*}[ht]
    \begin{center}
    \begin{tabular}{p{9cm} ccccc}
    \toprule
    & \multicolumn{2}{p{2cm}}{\centering \bf CBT NE} & \multicolumn{2}{p{2cm}}{\centering \bf CBT CN} \\
    & \bf Valid & \bf Test & \bf Valid & \bf Test\\
    \midrule
    Human \cite{hill-etal-2015} & - & {\em 81.6} & - & {\em 81.6} \\
    MemNN \cite{hill-etal-2015} & 70.4 & 66.6 & 64.2 & 63.0 \\ 
    AS Reader \cite{kadlec-etal-2016} & 73.8 & 68.6 & 68.8 & 63.4 \\
    GA Reader \cite{dhingra-etal-2016} & 74.9 & 69.0 & 69.0 & 63.9 \\
    Iterative Attention \cite{sordoni-etal-2016} & 75.2 & 68.6 & 72.1 & 69.2 \\
    AoA Reader \cite{cui-etal-2017-aoa} & 77.8 & 72.0 & 72.2 & 69.4 \\
    NSE Adp. Com. \cite{munkhdalai2016reasoning} & 78.2 & 73.2 & 74.2 & 71.4 \\
    GA Reader + Fine-gating \cite{yang-etal-2016} & 79.1 & {\em 75.0} & 75.3 & 72.0  \\
    AoA Reader + Re-ranking \cite{cui-etal-2017-aoa} & {\em 79.6} & 74.0 & {\em 75.7} & {\em 73.1} \\
    \midrule
    M-AoA Reader (GRU) & 78.0 & 73.8 & 72.8 & 69.8 \\
    M-AoA Reader (CRU) & 79.5 & 75.4 & 74.4 & 71.3 \\
    M-AoA Reader (CRU) + Re-ranking & {\bf 80.6} & {\bf 76.1} & {\bf 76.6} & {\bf 74.5} \\
    \midrule\midrule
    AS Reader (Ensemble) & 74.5 & 70.6 & 71.1 & 68.9 \\
    KnReader (Ensemble) & 78.0 & 73.3 & 72.2 & 70.6 \\
    Iterative Attention (Ensemble) & 76.9 & 72.0 & 74.1 & 71.0 \\
    AoA Reader (Ensemble) & 78.9 & 74.5 & 74.7 & 70.8 \\
    AoA Reader (Ensemble + Re-ranking) & {\em 80.3} & {\em 75.7} & {\em 77.0} & {\em 74.1} \\
    \midrule
    M-AoA Reader (CRU) (Ensemble) & 80.0 & 77.1 & 77.0 & 73.5 \\   
    M-AoA Reader (CRU) (Ensemble + Re-ranking) & {\bf 81.8} & {\bf 77.5} & {\bf 79.0} & {\bf 76.8} \\        
    \bottomrule
    \end{tabular}
    \end{center}
    \caption{\label{public-result} Results on the CBT NE and CN cloze-style reading comprehension datasets.
     }
    \end{table*}

\subsection{Experimental Setups}   
We also tested our CRU model in the cloze-style reading comprehension task.
We carried out experiments on the public datasets: CBT NE/CN \cite{hill-etal-2015}.
The CRU model used in these experiments is the {\em deep-enhanced} type with the convolutional filter length of 3.
In the re-ranking step, we also utilized three features: Global LM, Local LM, Word-class LM, as proposed by \citet{cui-etal-2017-aoa}, and all LMs are 8-gram trained by SRILM toolkit \cite{stolcke-2002}.
For other settings, such as hyperparameters, initializations, etc., we closely follow the experimental setups as \citet{cui-etal-2017-aoa} to make the experiments more comparable.

\subsection{Results}
The overall experimental results are given in Table \ref{public-result}. 
As we can see that our proposed models can substantially outperform various state-of-the-art systems by a large margin.

\begin{itemize}[leftmargin=*]
  \item Overall, our final model (M-AoA Reader + CRU + Re-ranking) could give significant improvements over the previous state-of-the-art systems by 2.1\% and 1.4\% in test sets, while re-ranking and ensemble bring further improvements.
  \item When comparing M-AoA Reader to the original AoA Reader, 1.8\% and 0.4\% improvements can be observed, suggesting that by incorporating additional features into embedding can enrich the power of word representation. Incorporating more additional features in the word embeddings would have another boost in the results, but we leave this in future work.
  \item Replacing GRU with our CRU could significantly improve the performance, where 1.6\% and 1.5\% gains can be obtained when compared to M-AoA Reader. This demonstrates that incorporating contextual information when modeling the sentence could enrich the representations. Also, when modeling an unknown word, except for its randomly initialized word embedding, the contextual information could give a possible guess of the unknown word, making the text more readable to the neural networks.
\item The re-ranking strategy is an effective approach in this task. We observed that the gains in the common noun category are significantly greater than the named entity. One possible reason is that the language model is much beneficial to CN than NE, because it is much more likely to meet a new named entity that is not covered in the training data than the common noun.
\end{itemize}

\section{Qualitative Analysis}\label{qualitative-analysis}        
In this section, we will give a qualitative analysis on our proposed CRU model in the sentiment classification task.
We focus on two categories of the movie reviews, which is quite harder for the model to judge the correct sentiment. The first one is the movie review that contains negation terms, such as ``not''. The second type is the one contains sentiment transition, such as ``clever but not compelling''. We manually select 50 samples of each category in the MR dataset, forming a total of 100 samples to see if our CRU model is superior in handling these movie reviews. The results are shown in Table \ref{quality-result-table}. As we can see that, our CRU model is better at both categories of movie review classification, demonstrating its effectiveness.

        \begin{table}[h]
        \begin{center}
        \begin{tabular}{lcc}
        \toprule
        & \bf GRU & \bf CRU \\
        \midrule
        Negation Term (50) & 37  & 42 \\
        Sentiment Transition (50) & 34  & 40 \\
        \midrule
        Total (100) & 71  & 82 \\
        \bottomrule
        \end{tabular}
        \end{center}
        \caption{\label{quality-result-table} Number of correctly classified samples.}
        \end{table}

Among these samples, we select an intuitive example that the CRU successfully captures the true meaning of the sentence and gives the correct sentiment label. We segment a full movie review into three sentences, which is shown in Table \ref{quality-table}.

        \begin{table}[htbp]
        \begin{center}
        \small
        \begin{tabular}{lcc}
        \toprule
        \bf Sentence & \bf GRU & \bf CRU \\
        \midrule
        I like that Smith & POS & POS \\
        \midrule
        I like that Smith, \\ he's {\em not making fun of} these people, & POS & POS \\
        \midrule
        I like that Smith, \\ he's {\em not making fun of} these people, \\he's {\em not laughing at} them. & {\em NEG} & POS \\
        \bottomrule
        \end{tabular}
        \end{center}
        \caption{\label{quality-table} Predictions of each level of the sentence.  }
        \end{table}

Regarding the first and second sentence, both models give correct sentiment prediction. While introducing the third sentence, the GRU baseline model failed to recognize this review as a positive sentiment because there are many negation terms in the sentence. However, our CRU model could capture the local context {\em during} the recurrent modeling the sentence, and the phrases such as ``not making fun'' and ``not laughing at'' could be correctly noted as positive sentiment which will correct the sentiment category of the full review, suggesting that our model is superior at modeling local context and gives much accurate meaning.

\section{Conclusion}\label{conclusion}
In this paper, we proposed an effective recurrent model for modeling sequences, called Contextual Recurrent Units (CRU). 
We inject the CNN into GRU, which aims to better model the local context information via CNN before recurrently modeling the sequence. 
We have tested our CRU model on the cloze-style reading comprehension task and sentiment classification task.
Experimental results show that our model could give substantial improvements over various state-of-the-art systems and set up new records on the respective public datasets. In the future, we plan to investigate convolutional filters that have dynamic lengths to adaptively capture the possible spans of its context.


\bibliography{conll2018}
\bibliographystyle{acl_natbib_nourl}

\end{document}